\documentclass[10pt, a4paper]{article}

\usepackage{lrec-coling2024} 

\usepackage{microtype}

\usepackage{inconsolata}

\usepackage{hyperref}       
\usepackage{url}            
\usepackage{booktabs}       
\usepackage{amsfonts}       
\usepackage{nicefrac}       
\usepackage{microtype}      
\usepackage{xcolor}         
\usepackage{standalone}
\usepackage{latexsym}
\usepackage{amsmath}
\usepackage{amssymb}
\usepackage{amsthm}
\usepackage{graphicx}
\usepackage{subcaption}
\usepackage{array}
\usepackage{tabu}
\usepackage{makecell}
\usepackage{paralist}
\usepackage{cases}
\usepackage{diagbox}
\usepackage{enumitem}
\usepackage{soul}
\usepackage{multirow}
\usepackage{verbatim}
\usepackage{tabulary}
\usepackage{booktabs}
\usepackage[mathscr]{euscript}
\usepackage{mathtools}
\usepackage{algorithm}
\usepackage{colortbl}
\usepackage{tabstackengine}
\usepackage{tikz}
\usepackage{relsize}
\usepackage{microtype}
\usepackage{tablefootnote}
\usepackage{silence}

\usepackage[noend]{algpseudocode}
\algnewcommand{\parState}[1]{\State%
    \parbox[t]{\dimexpr\linewidth-\algmargin}{\strut\hangindent=\algorithmicindent \hangafter=1 #1\strut}}

\algrenewcommand\algorithmicindent{1.0em}%

\usepackage{stmaryrd}
\usepackage{tikz-dependency}
\usetikzlibrary{automata,decorations.markings,arrows,positioning,matrix,calc,patterns,angles,quotes,calc}
\usepackage{adjustbox}
\usepackage{tabularx}
\usepackage{xspace}
\usepackage{tabulary}
\usepackage{afterpage}
\usepackage{hyperref}
\usepackage{url}
\usepackage{bm}
\usepackage{color}
\usepackage{graphicx}
\usepackage{slashbox}
\usepackage[toc,page]{appendix}
\usepackage{makecell}
\usepackage{boldline}

\definecolor{orange}{rgb}{1,0.5,0}
\definecolor{mdgreen}{rgb}{0.05,0.6,0.05}
\definecolor{mdblue}{rgb}{0,0,0.7}
\definecolor{dkblue}{rgb}{0,0,0.5}
\definecolor{dkgray}{rgb}{0.3,0.3,0.3}
\definecolor{slate}{rgb}{0.25,0.25,0.4}
\definecolor{gray}{rgb}{0.5,0.5,0.5}
\definecolor{ltgray}{rgb}{0.7,0.7,0.7}
\definecolor{purple}{rgb}{0.7,0,1.0}
\definecolor{lavender}{rgb}{0.65,0.55,1.0}

\definecolor{mypurple}{RGB}{111,61,121}
\definecolor{myblue}{RGB}{46,88,180}
\definecolor{myred}{RGB}{181,68,106}
\definecolor{myyellow}{RGB}{204,143,55}
\definecolor{mygray}{RGB}{128,128,128}
\definecolor{mygreen}{RGB}{126,198,54}

\newcommand{\interalia}[1]{\citep[\emph{inter alia}]{#1}}

\DeclareSymbolFont{extraup}{U}{zavm}{m}{n}
\DeclareMathSymbol{\vardiamond}{\mathalpha}{extraup}{87}

\newcolumntype{L}[1]{>{\raggedright\let\newline\\\arraybackslash\hspace{0pt}}m{#1}}
\newcolumntype{C}[1]{>{\centering\let\newline\\\arraybackslash\hspace{0pt}}m{#1}}
\newcolumntype{R}[1]{>{\raggedleft\let\newline\\\arraybackslash\hspace{0pt}}m{#1}}

\theoremstyle{definition}

\theoremstyle{remark}

\algrenewcommand{\algorithmiccomment}[1]{\leavevmode$\triangleright$ #1}

\setul{1pt}{.4pt}

\usepackage[shortcuts]{extdash}  

\usepackage{blindtext}
\usepackage{graphicx}
\usepackage{capt-of}
\usepackage{booktabs}
\usepackage{varwidth}
\newsavebox\tmpbox

\usepackage{amsmath,amsfonts,bm}

\def\eqref#1{equation~\ref{#1}}

\def\1{\bm{1}}

\def\rvy{{\mathbf{y}}}

\DeclareMathAlphabet{\mathsfit}{\encodingdefault}{\sfdefault}{m}{sl}
\SetMathAlphabet{\mathsfit}{bold}{\encodingdefault}{\sfdefault}{bx}{n}

\newcommand{\resolved}[1]{}

\newcommand{\com}[1]{}

\newcommand{\bos}{\textsc{bos}\xspace}
\newcommand{\eos}{\textsc{eos}\xspace}

\newcommand{\score}{\mathrm{score}}

\newcommand{\vocab}{\mathcal{V}}

\usepackage{algorithm}
\usepackage{colortbl}
\usepackage{tabstackengine}
\usepackage{tikz}
\usepackage{relsize}
\usepackage{microtype}
\DeclareMathSymbol{\varheart}{\mathalpha}{extraup}{86}

\DeclareRobustCommand{\greenbox}[1]{\setlength{\fboxsep}{1.0pt}\colorbox{green!25}{#1}}
\DeclareRobustCommand{\redbox}[1]{\setlength{\fboxsep}{1.0pt}\colorbox{red!15}{#1}}

\title{A Call for Clarity in Beam Search: How It Works and When It Stops}

\name{
\\
    \textbf{Jungo Kasai}$^{\heartsuit}$
\quad
\textbf{Keisuke Sakaguchi}$^{\clubsuit\square}$
\quad
\textbf{Ronan Le Bras}$^{\diamondsuit}$
\\
\textbf{Dragomir Radev}$^{\spadesuit}$
\quad
\textbf{Yejin Choi}$^{\varheart\diamondsuit}$
\quad
\textbf{Noah A.\ Smith}$^{\varheart\diamondsuit}$\\ 
\\
$^{\heartsuit}$Toyota Technological Institute at Chicago\quad $^{\clubsuit}$Tohoku University \quad $^{\square}$RIKEN\\
 $^{\diamondsuit}$Allen Institute for AI 
 \quad   $^{\spadesuit}$Department of Computer Science, Yale University
 \\
 $^{\varheart}$Paul G.\ Allen School of Computer Science \& Engineering, University of Washington
 \\
    {\tt jkasai@ttic.edu}
}

\address{}
\abstract{
Text generation with beam search has proven successful in a wide range of applications.
We point out that, though largely overlooked in the literature, the commonly-used implementation of beam decoding (e.g., Hugging Face Transformers and fairseq) uses a \textit{first come, first served} heuristic: it keeps a set of already completed sequences over time steps and stops when the size of this set reaches the beam size.
Based on this finding, we introduce a \textit{patience factor}, a simple modification to this beam decoding implementation, that generalizes the stopping criterion and provides flexibility to the depth of search.
Empirical results demonstrate that adjusting this patience factor improves decoding performance of strong pretrained models on news text summarization and machine translation over diverse language pairs, with a negligible inference slowdown. 
Our approach only modifies one line of code and can be thus readily incorporated in any implementation.\footnote{Our codebase is available at \url{anonymized}.}
Further, we find that different versions of beam decoding result in large performance differences in summarization, demonstrating the need for clarity in specifying the beam search implementation in research work.
Our code will be available upon publication.
 \\ \newline }

\begin{document}

\maketitleabstract

\section{Introduction}
Beam search has become a dominant inference algorithm for a wide range of language generation tasks, such as machine translation \cite{seq2seq,bahdanau2015attention,vaswani2017attention}, summarization \cite{nallapati-etal-2016-abstractive,see-etal-2017-get}, and image captioning \cite{Anderson2017up-down,li2020oscar}.
Beam decoding
is an approximate, pruned version of breadth-first search that seeks the highest-probability sequence under an autoregressive (left-to-right) language generation model.
In this work, we examine a popular implementation of beam decoding and propose a simple modification (one line of code) that improves the decoding performance of strong, neural language generation models (Fig.\ \ref{alg:fcfs}).

\begin{figure}[t!]
\algrenewcommand{\algorithmiccomment}[1]{\hskip1em\# \footnotesize#1 \small}
\vspace{-0.3cm}
\begin{algorithm}[H]
\small
\hspace*{1.5em} $k$: beam size, \ $M$: maximum length, \ $\vocab$: Vocabulary\ \\
\hspace*{1.5em} $\score(\cdot)$: scoring function, \greenbox{$p$: patience factor.}
\begin{algorithmic}[1]
\small \State $B_0 \gets \{ \langle 0, \bos \rangle \}$, $F_0  \gets \varnothing $
\small \For{ $t \in \{ 1, \dots, M\!-\! 1 \}$ }
    \small \State $H \gets \varnothing$, $F_t \gets F_{t-1}$
    \small \For{$ \langle s, \rvy \rangle \in B_{t-1}$} \algorithmiccomment{Expansion.}
    \small \For{$y \in \vocab$}
       \small  \State $s \gets \score(\rvy \circ y)$, \label{line:bs-score-eval}
\small         \ $H.\mathrm{add}( \langle s, \rvy \circ y \rangle)$
    \EndFor
    \EndFor
\small \State $B_t \gets \varnothing$
     \small \While{$ |B_t| < k$}  \algorithmiccomment{Find top $k$ w/o $\eos$ from $H$.}
    \small \State $ \langle s, \rvy \rangle \gets H.\mathrm{max}()$  
\small
\small     \If{\ $\rvy.\mathrm{last}() = \eos$}
\small        \State $F_t.\mathrm{add}(\langle s, \rvy \rangle )$    \algorithmiccomment{Finished hypotheses.}\label{line:fcfs_f}
    \small \Else{\ $B_t.\mathrm{add}(\langle s, \rvy \rangle)$}
     \EndIf
     \newline
  \small   \hspace*{-\fboxsep}\colorbox{green!25}{\parbox{1.0\linewidth}{\If{$|F_t|\geq k \cdot p$}\algorithmiccomment{Originally, $p\!=\!1$.}}}  \label{line:fcfs_p}
\small     \parbox{1.0\linewidth}{\State \ \ \ \ \Return$F_t.\mathrm{max}()$ \EndIf}
    \small  \State $H.\mathrm{remove}(\langle s, \rvy \rangle)$
\EndWhile
\EndFor
\State \Return $F_t.\mathrm{max}()$

\end{algorithmic}
\caption*{\small \textbf{FCFS Beam Decoding with Controlled Patience}}
\end{algorithm}
\vspace{-0.5cm}
\caption{First come, first served (FCFS) beam decoding with patience factor $p$.
The common implementation can be considered as a special case where $p\!=\!1$. The \greenbox{highlighted line} is the \textit{only} modification that this work introduces for performance improvement.
$F_t$: already completed sequences; $B_t$: beam of continuing sequences. 
$H_t$: expanded hypotheses before the top-$k$ operation. 
The input sequence to $\score$ is omitted.
}
\label{alg:fcfs}
\vspace{-0.6cm}
\end{figure}

We first bring attention to implementation variations of beam decoding that are largely ignored in the literature: a widely-used implementation of beam language decoding (e.g., \texttt{fairseq}, \citealp{ott2019fairseq}; Hugging Face's \texttt{Transformers}, \citealp{wolf-etal-2020-transformers})\footnote{\url{https://github.com/pytorch/fairseq/blob/main/fairseq/sequence_generator.py}; \url{https://github.com/huggingface/transformers/blob/master/src/transformers/generation_utils.py}.} follows a \textit{first come, first served} (FCFS) heuristic: when a total of $k$ finished candidates is found ($k$ is the beam size), it returns the best one from the $k$ candidates and discards all of the current, unfinished $k$ sequences in the beam.
Thus in practice, beam size $k$ determines both the breadth and depth of search.
We propose a \textit{patience factor} (Fig.\ \ref{alg:fcfs}) that decomposes these two roles and controls how many finished candidates have to be found before terminating the decoding.
The patience factor generalizes the commonly-used implementation and provides flexibility in the depth of beam search by changing the stopping criterion.

We apply the one-line modification to strong off-the-shelf transformer models without any change to the trained models for machine translation \cite{tang-etal-2021-multilingual} and text summarization \cite{lewis-etal-2020-bart}.
Our experiments demonstrate that the simple modification improves performance on the CNN/Dailymail \cite{cnndaily} and XSUM \cite{xsum2018} news summarization tasks and the WMT 2020/2021 machine translation tasks \cite{barrault-etal-2020-findings,akhbardeh-etal-2021-findings} across diverse language pairs.
Further, the introduction of the patience factor only results in a negligible inference slowdown, confirming its practical advantage in downstream applications.

Our analysis shows that, while the performance gain is sensitive to hyperparameters of beam decoding (beam size and length penalty; \citealp{johnson-etal-2017-googles}), the patience factor is consistently beneficial.
Moreover, we extensively compare our results with the \textit{vanilla} implementation of beam search that much prior work assumes \interalia{meister-etal-2020-best,stahlberg-byrne-2019-nmt}, though it is not used in popular libraries in practice.
Empirically, we found that the vanilla algorithm performs competitively with FCFS on machine translation but \textbf{substantially underperforms on summarization}.
Therefore, although much prior work on text decoding implicitly assumes the vanilla version of beam decoding, researchers should specify which implementation is used when reporting results.


\section{A Call for Clarity in Beam Decoding}\label{sec:beam}
\begin{figure}[t!]
\vspace{-0.3cm}
\begin{algorithm}[H]
\algrenewcommand{\algorithmiccomment}[1]{\hskip1em$\#$ \footnotesize#1 \normalsize}
\small
\hspace*{2.5em} $k$: beam size, \ $M$: maximum length, \\
\hspace*{2.5em} $\vocab$: Vocabulary, \
 $\score(\cdot)$: scoring function.
\begin{algorithmic}[1]
\State $B_0 \gets \{ \langle 0, \bos \rangle \}$
\For{ $t \in \{ 1, \dots, M\!-\! 1 \}$ }
    \small \For{$ \langle s, \rvy \rangle \in B_{t-1}$}
     \small \If{\ $\rvy.\mathrm{last}() = \eos$}
     
      \small \State \colorbox{red!15}{$H.\mathrm{add}(\langle s, \rvy \rangle )$} \label{line:vanilla_h}
      \small \State \textbf{continue}
    \EndIf
    \small \For{$y \in \vocab$}
        \small \State $s \gets \score(\rvy \circ y)$,
        \ $H.\mathrm{add}( \langle s, \rvy \circ y \rangle)$
    \EndFor
    \EndFor
    \small \State  $B_t \gets \varnothing$
\small \While{$ |B_t| < k $} \algorithmiccomment{Find top $k$ from $H$.}
\small \State $ \langle s, \rvy \rangle \gets H.\mathrm{max}()$, $B_t.\mathrm{add}(\langle s, \rvy \rangle)$
\small \State $H.\mathrm{remove}(\langle s, \rvy \rangle)$
\EndWhile

 \small \hspace*{-\algorithmicindent} \hspace*{-\algorithmicindent} \hspace*{-2\fboxsep}\colorbox{green!25}{\parbox{1.0\linewidth}{\If{$\rvy.\mathrm{last}() = \eos$, $\forall \rvy \in B_t$}\algorithmiccomment{All finished.}\EndIf}} \label{line:vanilla_stop} 
 \vspace*{-0.3cm}
\small \State  \hspace*{\algorithmicindent} \Return $B_t.\mathrm{max}()$
\EndFor
\small \State \Return $B_t.\mathrm{max}()$

\end{algorithmic}
\caption*{\small \textbf{Vanilla Beam Decoding}}
\end{algorithm}
\vspace{-0.5cm}
\caption{The vanilla version of beam decoding that much prior work assumes \textbf{departs from popular libraries, such as Hugging Face \texttt{Transformers}}. The top-$k$ operation is applied over $H$, the union of the finished and continuing sequences.
This is implemented, for example, in the TensorFlow Addons library \cite{tensorflow2015-whitepaper}.\protect\footnotemark{}
See also \citet{stahlberg-byrne-2019-nmt,meister-etal-2020-best}.
}
\label{alg:vanilla}
\vspace{-0.2cm}
\end{figure}
\footnotetext{\url{https://www.tensorflow.org/addons/api_docs/python/tfa/seq2seq/BeamSearchDecoder}.}

\begin{figure}[h!]
\centering
    \includegraphics[width=0.45\textwidth]{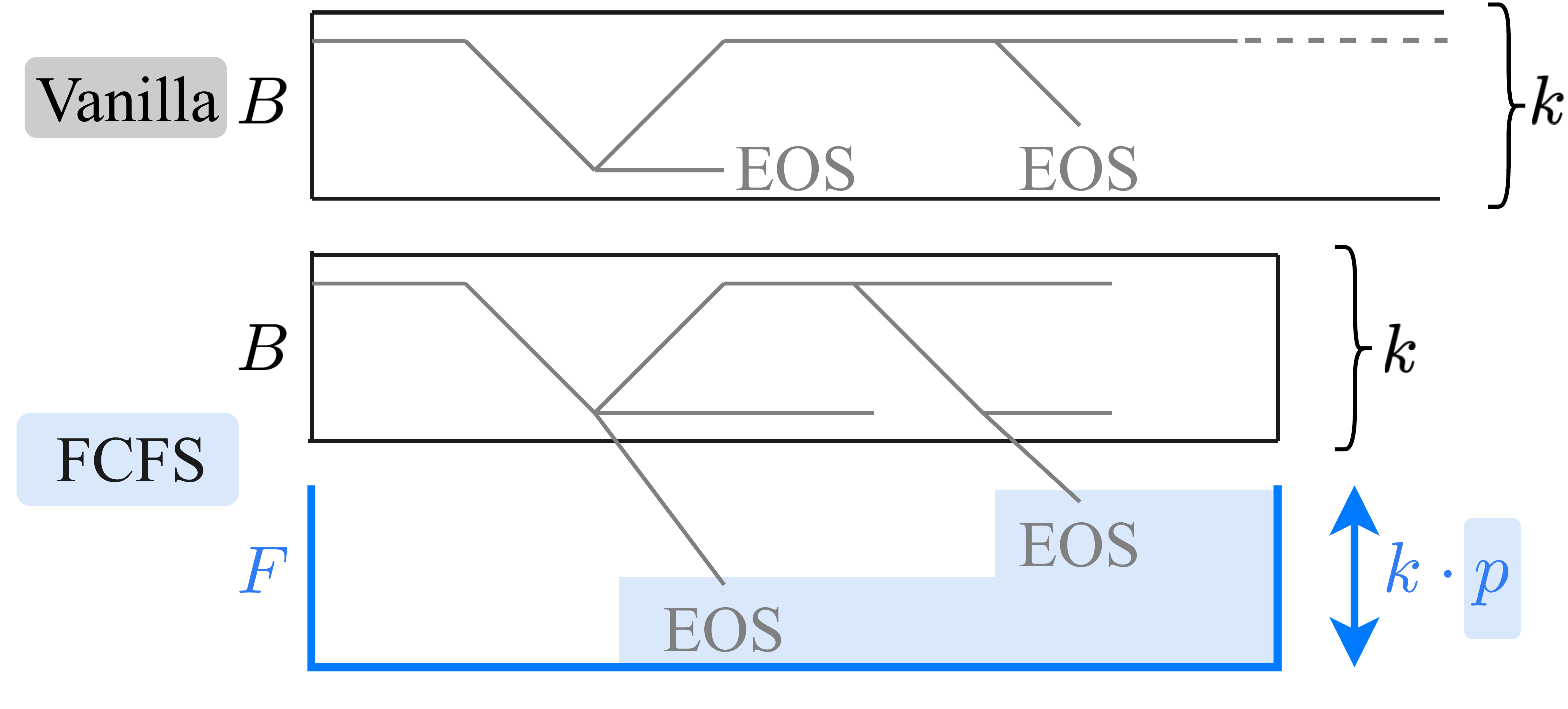}
\caption{FCFS with patience factor $p$ vs.\ vanilla beam decoding. $k$ denotes the beam size. FCFS stores finished sentences in $F$, but they stay in (and later may fall off from) beam $B$ during vanilla decoding. \textbf{$k\! \cdot\! p$} determines the size of $F$.
The illustration of beam decoding here is inspired by \citet{huang-etal-2012-structured}.
}
\label{fig:constrast}
\vspace{-0.6cm}
\end{figure}
\textbf{Vanilla and FCFS Implementations \ }
Beam decoding has been applied to sequence-to-sequence models \cite{Graves2012SequenceTW,BoulangerLewandowski2013AudioCR,BoulangerLewandowski2013HighdimensionalST}, and it is now used in many state-of-the-art systems for language generation tasks \interalia{zhang2019pegasus,zhang2021vinvl,tran2021facebook,2020t5}.
\textbf{While largely ignored in the literature}, beam decoding has two major variations (Figs.\ \ref{alg:fcfs} and \ref{alg:vanilla}).
They differ primarily in the treatment of finished sequences with the $\eos$ symbol at the end: FCFS collects finished sequences in a \textit{first come, first served manner} and removes them from the beam (Line \ref{line:fcfs_f}, Fig.\ \ref{alg:fcfs}), whereas the vanilla version finds the top $k$ sequences, including both finished and unfinished sequences (\redbox{Line \ref{line:vanilla_h}} in Fig.\ \ref{alg:vanilla}). 
Our experiments will show that this difference can affect the downstream performance substantially, especially on summarization (\S\ref{sec:results}).


Further comparing Figs.\ \ref{alg:fcfs} and \ref{alg:vanilla}, we see their difference in terms of the breadth and depth of search.
Given the same beam size $k$, FCFS has a wider breadth since it collects $k$ unfinished sequences at every step regardless of how many sequences are finished with the $\eos$ symbol.\footnote{In practice, this is implemented by taking the top $2k$ sequences at every step. We find at most $k$ $\eos$ symbols, so there are always at least $k$ \textit{unfinished} sequences.
See \url{https://github.com/huggingface/transformers/}.}
The vanilla algorithm decodes until all top-$k$ sequences are finished (\colorbox{green!25}{Line \ref{line:vanilla_stop}}, Fig.\ \ref{alg:vanilla}), and therefore it tends to result in deeper search.
FCFS, in contrast, terminates when a total of $k$ finished sequences is found.


\noindent \textbf{Patience Factor for FCFS \ }
Beam size $k$ in FCFS thus controls both the breadth and stopping criterion (i.e., depth) of search.
We introduce the patience factor (\greenbox{Line \ref{line:fcfs_p}}, Fig.\ \ref{alg:fcfs}) that relaxes this assumption and separates the stopping criterion from the search breadth.
Fig.\ \ref{fig:constrast} illustrates this patience factor as well as the difference between the FCFS and vanilla algorithms.
The one-line change generalizes FCFS ($p\!=\!1$) and adds flexibility.
We will show that this flexibility is beneficial, espeically on news text summarization (\S\ref{sec:results}).

\begin{table*}[t!]
\addtolength{\tabcolsep}{-1.0pt}  
\centering
\small
\renewcommand{\arraystretch}{1.3}
\begin{tabular}{l cc cc  cc cc m{0.5em} ccc ccc }
\toprule
&\multicolumn{8}{c}{WMT 2020/2021 Machine Translation ($p\!=\!2$)} & 
& \multicolumn{6}{c}{Summarization ($p\!=\!0.5$)} \\
\cmidrule(lr){2-9}
\cmidrule(lr){11-16}
&\multicolumn{2}{c}{\textbf{EN$\leftrightarrow$DE}} &  \multicolumn{2}{c}{\textbf{EN$\leftrightarrow$JA}} & \multicolumn{2}{c}{\textbf{EN$\leftrightarrow$PL}}  & \multicolumn{2}{c}{\textbf{EN$\leftrightarrow$ZH}} &
& \multicolumn{3}{c}{\textbf{CNNDM}} & 
\multicolumn{3}{c}{\textbf{XSUM}}\\
\cmidrule(lr){2-3}
\cmidrule(lr){4-5}
\cmidrule(lr){6-7}
\cmidrule(lr){8-9}
\cmidrule(lr){11-13}
\cmidrule(lr){14-16}
\textbf{Algorithm} & $\rightarrow$ & $\leftarrow$&   $\rightarrow$ & $\leftarrow$  & $\rightarrow$ & $\leftarrow$  & $\rightarrow$ & $\leftarrow$ &
& R-2 & R-3 & R-L & R-2 & R-3 &R-L\\
 \hline
Greedy & 43.7& 66.2 & 33.6 & 9.5 & 46.0 & 53.5 & 32.5 & 23.5 &
& 21.1 & 11.9 & 30.7 & 19.8  & 10.7& 34.3\\
Vanilla & 48.2 & 66.3 & \textbf{38.7} & \textbf{15.7} & 52.7 & 58.2 & \textbf{33.9} & 29.9 &
& 19.2 & 11.0 & 28.0 & 19.5 &10.7 & 33.1 \\
FCFS & 47.9 & 66.2 & 38.0 & 15.0 & 52.1 & 58.1 & 33.7 & 29.6 &
& 20.4 & 11.6 &  30.3 & 20.4 & 11.4 & 34.4\\
FCFS w/ $p$ & \textbf{48.3} & \textbf{66.4} & 38.4& 15.6& \textbf{53.0} & \textbf{58.4} & 33.8 & \textbf{30.2} & 
& \textbf{21.4}& \textbf{12.4} &\textbf{31.2} & \textbf{21.0} &\textbf{11.8} & \textbf{35.4} \\
\bottomrule
\end{tabular}
\caption{We evaluate three beam decoding variations, as well as greedy decoding, on the machine translation and summarization test data with the COMET score \cite{comet-wmt} and ROUGE scores (ROUGE-2/3/L), respectively. 
FCFS w/ p indicates our FCSF algorithm with the patience factor ($p\!=\!2$ for machine translation and $p\!=\!0.5$ for summarization).
For CNNDM, we used 100 test articles with 10 human-written references \cite{kryscinski-etal-2019-neural}.
}
\vspace{-0.5cm}
\label{tab:results}
\end{table*}
\section{Experiments}\label{sec:experiments}
We compare implementation variations of beam decoding on text summarization and machine translation over a wide range of language pairs.

\subsection{Experimental Setup for Beam Decoding}
In addition to \textbf{greedy search}, we evaluate three beam decoding variations on machine translation and summarization: \textbf{vanilla}, \textbf{FCFS}, and \textbf{FCFS with the patience factor}.\footnote{Other beam decoding variants target various purposes, like output diversity (stochastic beam search; \citealp{kool19,meister-etal-2021-conditional}) or efficiency (best-first beam search; \citealp{meister-etal-2020-best}). Here we focus on the commonly-adopted implementations and the top-1 output quality.}
For machine translation, we use multilingual BART \cite{tang-etal-2021-multilingual}, a strong, pretrained transformer model,\footnote{\url{https://github.com/pytorch/fairseq/tree/main/examples/multilingual\#mbart50-models}.} and WMT 2020/2021 news test data \cite{barrault-etal-2020-findings,akhbardeh-etal-2021-findings} for four diverse language pairs (eight directions): WMT 2020 for EN$\leftrightarrow$PL (Polish) and 2021 for EN$\leftrightarrow$DE (German),  EN$\leftrightarrow$JA (Japanese), and  EN$\leftrightarrow$ZH (Chinese).
We apply beam decoding with the same hyperparameters as \citet{tang-etal-2021-multilingual}: beam size 5 and length penalty 1.
We measure performance with the COMET score \cite{rei-etal-2020-comet,comet-wmt}, a state-of-the-art evaluation metric based on multilingual contextual representations.
For summarization, we experiment with the CNN/Dailymail (CNNDM, \citealp{cnndaily}) and  XSUM \cite{xsum2018} datasets.
We apply the off-the-shelf BART models \cite{lewis-etal-2020-bart} that are finetuned on each dataset.\footnote{\url{https://github.com/pytorch/fairseq/tree/main/examples/bart}.}
Performance is measured with ROUGE scores \cite{Lin2004ROUGEAP}.
We follow the original setting in \citet{lewis-etal-2020-bart}: beam sizes 4 and 6 and length penalty 2 and 1 for CNNDM and XSUM, respectively. 
More experimental details are described in the appendix (following the submission guideline, the appendix will be added in the final version).

We experiment with the same patience factor on all datasets for each task, based on our preliminary development: $p\!=\!2$ for machine translation and $p\!=\!0.5$ for summarization.
Here we avoid additional effort and demonstrate the practical value of our simple modification.
We present detailed sensitivity analysis over $p$ in \S\ref{sec:analysis}.
Note that a larger $p$ implies deeper search, but deeper (and thus more accurate) search does not necessarily result in better generations, similar to the \textit{beam search curse} \cite{koehn-knowles-2017-six}.

\subsection{Results}
\label{sec:results}
Seen in Table \ref{tab:results} are results from our experiments.
FCFS with the patience factor outperforms the widely-used FCFS variation across the board; e.g., 53.0 vs.\ 52.1 on EN$\rightarrow$PL.
Particularly noteworthy are the performance gains on the two summarization datasets; e.g., 31.2 vs.\ 30.3 ROUGE-L on CNNDM.
Comparing vanilla decoding and FCFS, we see that the former outperforms the latter (and is competitive with or slightly better than FCFS w/ $p$) on machine translation but underperforms substantially on summarization; e.g., 34.4 vs.\ 33.1 ROUGE-L on XSUM.
Vanilla beam decoding even performs worse than greedy decoding in many cases. 
This large degradation from the vanilla implementation illustrates \textbf{the importance of specifying the version of beam decoding}, which is not commonly done in practice.

\subsection{Analysis}
\label{sec:analysis}
\begin{figure}[h!]
\vspace{-0.2cm}
\centering
    \includegraphics[width=0.48\textwidth]{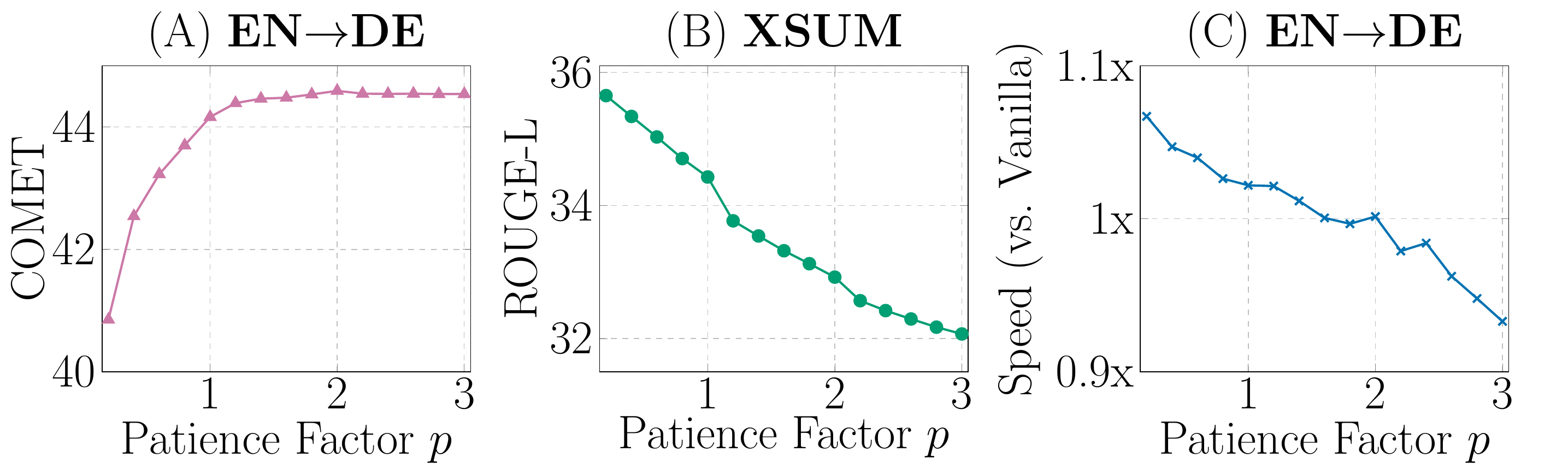}
\caption{Effects of varying patience factors $p$ on the dev.\ score (A and B) and inference speed (C). The inference speed is measured with batch size 20, relative to the vanilla version on the same single Nvidia A100-SXM GPU. Other languages pairs were similar to EN$\rightarrow$DE (A). CNNDM also had similar trends to XSUM (B).}
\label{fig:varying_p}
\vspace{-0.5cm}
\end{figure}

Here we use the standard dev.\ split from the XSUM dataset and news test 2020 EN$\rightarrow$DE and ZH$\rightarrow$EN data.
We fixed the value of $p$ for each task so far, but Fig.\ \ref{fig:varying_p} explores varying patience factors and their effects on the performance (A: EN$\rightarrow$DE; B: XSUM) and the inference speed (C).
The translation performance improves with larger patience factors with diminishing gains.
Surprisingly, summarization benefits from patience factors smaller than the original value of 1, possibly due to the nature of the summarization task that aims to generate concise text.
Note that this observation does \emph{not} contradict with prior work because the search accuracy of beam decoding does \emph{not} always correlate with the output quality \interalia{koehn-knowles-2017-six,stahlberg-byrne-2019-nmt,meister-etal-2020-beam}.
Regardless, our patience factor provides useful flexibility for any generation task. 

\begin{figure}[h!]
\centering
    \includegraphics[width=0.48\textwidth]{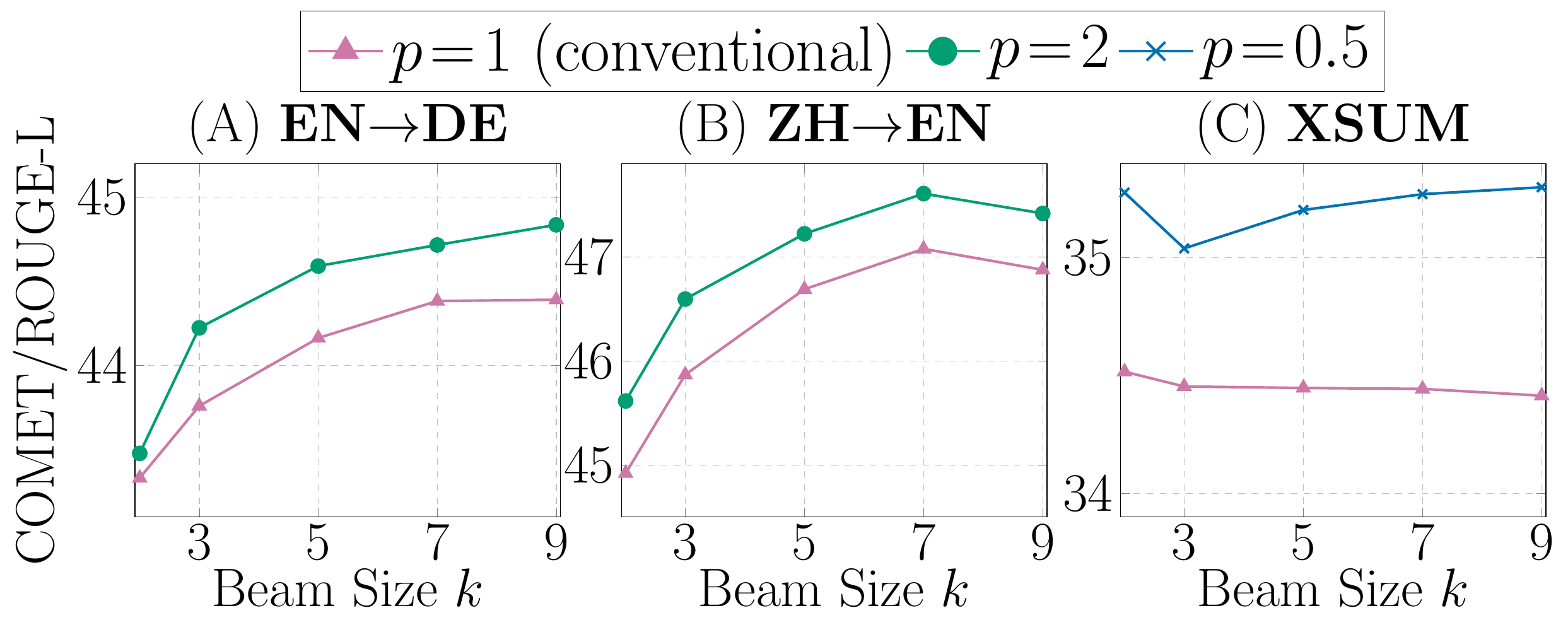}
\caption{Effects of controlled patience on the dev.\ data over varying beam sizes.
The length penalty value is 1.
We evaluate with COMET for machine translation and ROUGE-L for XSUM summarization.}
\label{fig:varying_beam}
\vspace{-0.5cm}
\end{figure}

As expected, generation slows down as $p$ increases (Fig.\ \ref{fig:varying_p}C). The inference slowdown from around $p\!=\!2$ is still negligible, again showing the practicality of our method.
Fig.\ \ref{fig:varying_beam} explores the performance gains from the patience factor over varying beam sizes.
The amount of improvement changes, but the patience factor is generally beneficial.
We see similar trends for various values of the length penalty (see the appendix).


\section{Further Related Work}\label{sec:related}
\textbf{Stopping Criteria for Beam Decoding \ }
The patience factor changes the stopping criterion and adds flexibility in the search depth of the widespread variation of beam decoding.
Similarly, several prior works studied stopping criteria to improve machine translation \cite{huang-etal-2017-finish,yang-etal-2018-breaking,ma-etal-2019-learning}.
Our experiments are consistent with their findings: stopping criteria that yield deeper search can improve the machine translation performance.

\noindent \textbf{Breadth of Beam Decoding \ }
Much prior work explored downstream effects of the search breadth \interalia{koehn-knowles-2017-six,murray-chiang-2018-correcting,ott2018icml,pmlr-v97-cohen19a,stahlberg-byrne-2019-nmt}.
Beam decoding with larger beam sizes can find sequences with higher scores but lead to performance degradation (often called the \textit{beam search curse}; \citealp{yang-etal-2018-breaking}).
Recent work \cite{meister-etal-2020-beam} argued that beam decoding with small beams introduces bias that is related to the uniform information density of human-produced text \cite{levythesis}.
\citet{freitag-al-onaizan-2017-beam} proposed a method to adaptively shrink the beam width based on the partial scores to speed up inference.
We focused on the stopping criteria (i.e., depth) and separated them from the breadth of the common implementation.

\section{Conclusion}\label{sec:conclusion}
We brought attention to the crucial yet overlooked implementation differences of beam search.
We then introduced the patience factor for the widespread implementation.
Our experiments showed that the patience factor improves the generation performance especially on summarization, with an insignificant slowdown in generation.
As it only requires a minimal change in code, we hope that many researchers and practitioners of language generation will benefit from our simple yet effective modification.
\section*{Limitations}
We evaluated our decoding method both on machine translation and news summarization.
Our machine translation experiments span diverse languages, including morphologically rich languages (e.g., Japanese and Polish) and languages with non-Latin scripts (e.g., Japanese and Chinese). 
Nonetheless, our summarization experiments are limited to English and the news domain mainly due to our budget constraints.
There are also many other language generation tasks for which our method can be useful.
Since our improvement only requires one line of code, we hope that practitioners will implement it for the domain and the task of their interest and further assess how our decoding algorithm performs over a wider range of applications.

Evaluating language generation remains a challenging research problem.
We carefully set up our experiments to mitigate potential evaluation issues.
The WMT 2020/2021 test data consist only of news text written in the original language, in contrast to the test data from WMT 2018 \cite{bojar-etal-2018-findings} or earlier.
For example, the WMT 2021 EN$\rightarrow$DE and DE$\rightarrow$EN test data come from completely different documents.
This avoids the \textit{translationese effect} that would overestimate the translation performance due to the simplicity of translated text \cite{Graham2020TranslationeseIM}.
Moreover, some language pairs in the WMT 2020 and 2021 test data have multiple references per instance, which increases the correlation of automatic evaluations with human judgment \cite{billboard}.
We presented results using automatic metrics from recent work \cite{comet-wmt} as well as conventional, n-gram overlap metrics \cite{Papineni2001BleuAM,Lin2004ROUGEAP}.
Recent automatic metrics have shown to have higher correlation with human judgements, but human judgments are sometimes inconsistent, especially when crowdsourced \cite{clark-etal-2021-thats,kasai2021thumb}.
Since our decoding method is a generalization of the widely-used beam search algorithm, we hope that it will be tested and used in real-world systems of language generation.

\nocite{*}

\bibliographystyle{lrec-coling2024-natbib}
\bibliography{lrec-coling2024-example}


\bibliographystylelanguageresource{lrec-coling2024-natbib}
\bibliographylanguageresource{languageresource}

\appendix
\begin{appendices}
\begin{table}[h]
\addtolength{\tabcolsep}{10.0pt}  
\small
\centering
\begin{tabular}{@{} l r @{}}
\toprule[.1em]
\textbf{Hyperparameter} & \textbf{Value}\\
\midrule[.1em]
\multicolumn{2}{c}{\textbf{WMT Machine Translation (All Pairs)}}\\
beam size & 5 \\
length penalty & 1\\
\midrule[.05em]
\multicolumn{2}{c}{\textbf{CNNDM Summarization}}\\
beam size & 4 \\
length penalty & 2\\
max-len-b & 140\\
min-len & 55\\
no-repeat-ngram-size &  3 \\
\midrule[.05em]
\multicolumn{2}{c}{\textbf{XSUM Summarization}}\\
beam size & 6 \\
length penalty & 1\\
max-len-b & 60\\
min-len & 10\\
no-repeat-ngram-size &  3 \\

\bottomrule[.1em]
\end{tabular}
\caption{Beam decoding hyperparameters. We generally followed prior work: \citet{tang-etal-2021-multilingual} for machine translation and \citet{lewis-etal-2020-bart} for CNNDM and XSUM summarization.}
\label{tab:hyper}
\end{table}

\begin{table*}[t!]
\addtolength{\tabcolsep}{0.0pt}  
\centering
\small
\renewcommand{\arraystretch}{1.3}
\begin{tabular}{ lcc cc cc cc m{0.1em} c c }
\toprule
&\multicolumn{8}{c}{WMT 2020 and 2021 Machine Translation (BLEU)} & 
& \multicolumn{2}{c}{Summarization} \\
\cmidrule(lr){2-9}
\cmidrule(lr){11-12}
&\multicolumn{2}{c}{\textbf{EN$\leftrightarrow$DE}} &  \multicolumn{2}{c}{\textbf{EN$\leftrightarrow$JA}} & \multicolumn{2}{c}{\textbf{EN$\leftrightarrow$PL}}  & \multicolumn{2}{c}{\textbf{EN$\leftrightarrow$ZH}} &
& \multicolumn{1}{c}{\textbf{CNNDM}} & 
\multicolumn{1}{c}{\textbf{XSUM}}\\
\cmidrule(lr){2-3}
\cmidrule(lr){4-5}
\cmidrule(lr){6-7}
\cmidrule(lr){8-9}
\cmidrule(lr){11-11}
\cmidrule(lr){12-12}
\textbf{Algorithm} & $\rightarrow$ & $\leftarrow$&   $\rightarrow$ & $\leftarrow$  & $\rightarrow$ & $\leftarrow$  & $\rightarrow$ & $\leftarrow$ &
& COMET & COMET\\
 \hline
Greedy & 42.9& 46.6 & 20.2 & 17.4 & 19.8 & 30.7 & 31.2 & 21.7 &
& \textbf{1.6} & 0.1\\
Vanilla & \textbf{45.1} & 48.4 & 21.6 & 19.7 & \textbf{21.1} & \textbf{32.5} & \textbf{32.5} & 23.6 &
& -5.5 & -1.6\\
FCFS & 45.0 & 48.4 & 21.3 & 19.5 & 21.0 & 32.4 & \textbf{32.6} & 23.4 &
& -4.2 & 2.2\\
FCFS w/ $p$ & 45.0 & \textbf{48.5} & \textbf{21.7}& \textbf{19.8} & \textbf{21.1} & \textbf{32.5} & 32.3 & \textbf{23.7} & 
& -1.1 & \textbf{2.5}\\
\bottomrule
\end{tabular}
\caption{We evaluate three beam decoding variations, as well as greedy decoding, on machine translation and summarization and report BLEU \cite{Papineni2001BleuAM} and COMET \cite{comet-wmt} scores here.
FCFS w/ $p$ indicates our FCSF algorithm with the patience factor ($p\!=\!2$ for machine translation and $p\!=\!0.5$ for summarization).
COMET is an automatic metric for machine translation that uses crosslingual contextual representations from XLM-RoBERTa \cite{conneau-etal-2020-unsupervised}, but it can be used for evaluating summarization as well \cite{billboard}.}
\label{tab:additional_results}
\end{table*}

\section{Beam Decoding Hyperparameters}
\label{sec:hyper}
Table \ref{tab:hyper} shows the beam decoding hyperparameters in our experiments. We generally follow the original settings of the pretrained, off-the-shelf models \cite{tang-etal-2021-multilingual,lewis-etal-2020-bart}. 

\section{Additional Results}
\label{sec:additional_results}
Table \ref{tab:additional_results} reports BLEU \cite{Papineni2001BleuAM} and COMET \cite{comet-wmt} scores for the machine translation and summarization experiments, respectively.
We use the \texttt{sacreBLEU} implementation for BLEU \cite{post-2018-call}.
Note that much recent work \interalia{tangledup,billboard,kasai2021thumb,Edunov2020OnTE} found poor correlation between BLEU scores and human judgment for evaluating strong language generation models.
COMET is an automatic metric for machine translation that uses crosslingual contextual representations from XLM-RoBERTa \cite{conneau-etal-2020-unsupervised}, but it can be used \textit{monolingually} for evaluating summarization as well \cite{billboard}.

\begin{figure}[h!]
\centering
    \includegraphics[width=0.499\textwidth]{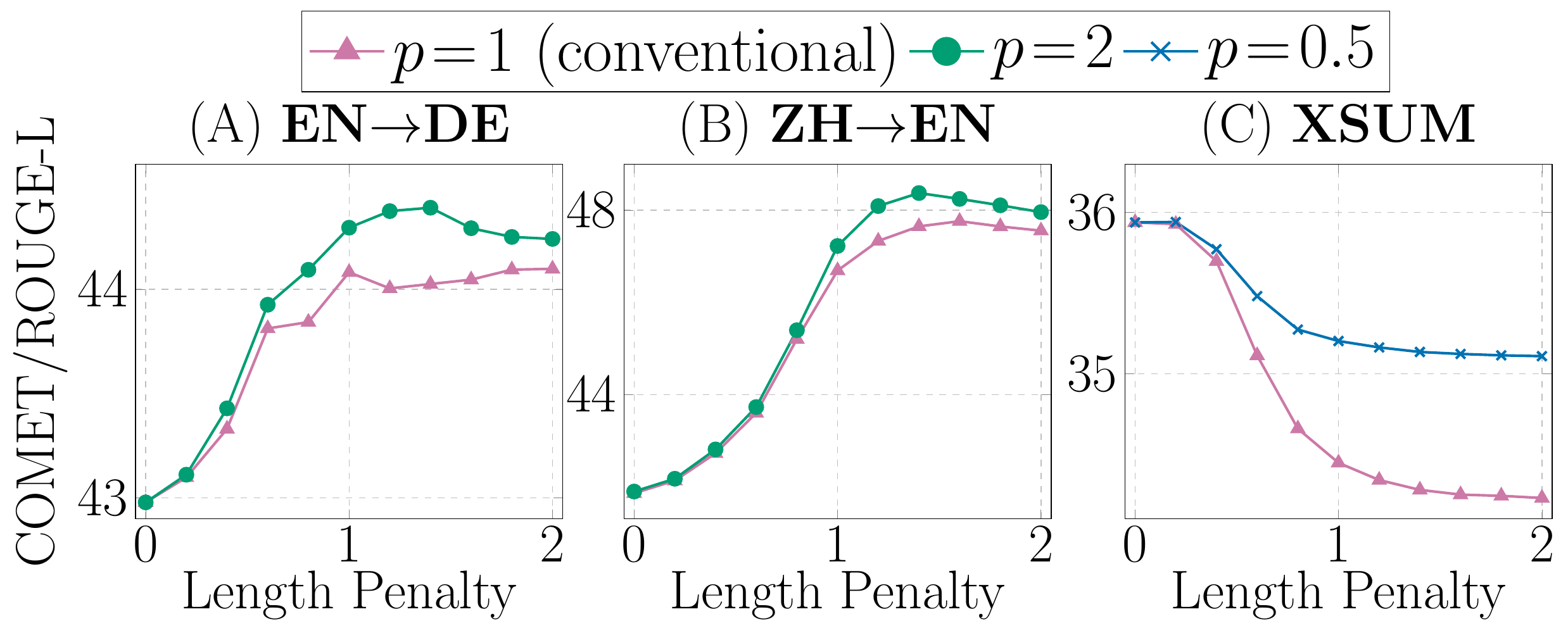}
\caption{Effects of controlled patience on the dev.\ data over varying length penalty values.
The beam sizes are all 5.
We evaluate with COMET for machine translation and ROUGE-L for XSUM summarization.}
\label{fig:varying_lenpen}
\vspace{-0.0cm}
\end{figure}

Fig.\ \ref{fig:varying_lenpen} explores the performance gains from the patience factor over varying length penalty values.
Consistent with the trends from various beam sizes (Fig.\ \ref{fig:varying_beam}), the amount of improvement changes, but the patience factor is generally beneficial.

\end{appendices}

\end{document}